\documentclass[10pt,journal,compsoc]{IEEEtran}
\ifCLASSOPTIONcompsoc
  \usepackage[nocompress]{cite}
\else
  \usepackage{cite}
\fi

\usepackage{ragged2e} 
\usepackage[pdftex]{graphicx}
\usepackage{amsmath} 
\usepackage[ruled,linesnumbered]{algorithm2e}
\usepackage{booktabs}
\usepackage{multirow}
\usepackage{subfigure}
\usepackage{color}
\usepackage{xspace}

\makeatletter
\DeclareRobustCommand\onedot{\futurelet\@let@token\@onedot}
\def\@onedot{\ifx\@let@token.\else.\null\fi\xspace}
\def\eg{\emph{e.g}\onedot} 
\def\ie{\emph{i.e}\onedot} 
 
\def\etc{\emph{etc}\onedot} 
\def\wrt{w.r.t\onedot} 

\makeatother

\def\argmin{\operatornamewithlimits{arg\,min}}
\def\argmax{\operatornamewithlimits{arg\,max}}

\begin{document}
%
\title{Adversarial Rain Attack and Defensive Deraining for DNN Perception}
%
%
%
%

\author{Liming~Zhai,
        Felix~Juefei-Xu,~\IEEEmembership{Member,~IEEE,}
        Qing~Guo,~\IEEEmembership{Member,~IEEE,}
        Xiaofei~Xie,~\IEEEmembership{Member,~IEEE,}
        Lei~Ma,~\IEEEmembership{Member,~IEEE,}
        Wei~Feng,~\IEEEmembership{Member,~IEEE,}
        Shengchao~Qin,~\IEEEmembership{Senior~Member,~IEEE,}
        and~Yang~Liu,~\IEEEmembership{Senior~Member,~IEEE}
\IEEEcompsocitemizethanks{\IEEEcompsocthanksitem Liming Zhai, Qing Guo and Yang Liu are with Nanyang Technological University, 639798, Singapore.\protect\\
\IEEEcompsocthanksitem Felix Juefei-Xu is with Alibaba Group, Sunnyvale, CA 94085, USA.\protect\\
\IEEEcompsocthanksitem Xiaofei Xie is with Singapore Management University, 188065, Singapore.\protect\\
\IEEEcompsocthanksitem Lei Ma is with University of Alberta, Edmonton, AB T6G 2R3, Canada and also with Kyushu University, Fukuoka 819-0395, Japan.\protect\\
\IEEEcompsocthanksitem Wei Feng is with the School of
Computer Science and Technology, Tianjin
University, Tianjin 300305, China.\protect\\
\IEEEcompsocthanksitem Shengchao~Qin is with the School of Computing, Media and the Arts, Teesside University, Middlesbrough TS1 3BX, United Kingdom.}
}

%
%

\markboth{}%
{Shell \MakeLowercase{\textit{et al.}}: Bare Demo of IEEEtran.cls for Computer Society Journals}
%



\IEEEtitleabstractindextext{%
\begin{abstract}
\justifying
Rain often poses inevitable threats to deep neural network (DNN) based perception systems, and a comprehensive investigation of the potential risks of the rain to DNNs is of great importance. However, it is rather difficult to collect or synthesize rainy images that can represent all rain situations that would possibly occur in the real world.
To this end, in this paper, we start from a new perspective and propose to combine two totally different studies, \ie, rainy image synthesis and adversarial attack.
We first present an \textit{adversarial rain attack}, with which we could simulate various rain situations with the guidance of deployed DNNs and reveal the potential threat factors that can be brought by rain.
In particular, we design a \textit{factor-aware rain generation} that synthesizes rain streaks according to the camera exposure process and models the learnable rain factors for adversarial attack. 
With this generator, we perform the \textit{adversarial rain attack} against the image classification and object detection.
To defend the DNNs from the negative rain effect, we also present a defensive deraining strategy, for which we design an adversarial rain augmentation that uses mixed adversarial rain layers to enhance deraining models for downstream DNN perception.
Our large-scale evaluation on various datasets demonstrates that our synthesized rainy images with realistic appearances not only exhibit strong adversarial capability against DNNs, but also boost the deraining models for defensive purposes, building the foundation for further rain-robust perception studies.
\end{abstract}

}

\maketitle

\IEEEdisplaynontitleabstractindextext

%
\IEEEpeerreviewmaketitle

\IEEEraisesectionheading{\section{Introduction}\label{sec:introduction}}

%
%
%
%
\IEEEPARstart{R}{ain} is condensed aqueous vapor in the form of falling drops with high speeds and small sizes. As a common weather phenomenon, rain can bring massive impacts not only on our human society, but also significant influences on today’s intelligent era. Such impacts are most prominent in deep neural network (DNN) based perception systems, \eg, autonomous driving, video surveillance and unmanned aerial vehicle (UAV), which can be easily disturbed by the inevitable rain effects, suffering from severe safety and security issues \cite{zhang2018deeproad,bahnsen2018rain}. Therefore, it is of great importance and pressing to comprehensively study how the rain affects the DNNs and how to reduce the negative rain effect.

The rain effect in images to DNNs has something in common with adversarial examples, which are typically generated by adding well-crafted perturbations to benign samples under some constraints \cite{yuan2019adversarial}. In particular, rain can also mislead the DNNs as if by adding minor rain noises (\ie, some non-trivial perturbation) to clean images. However, the reasons behind the adversarial noise attack and rain effect are obviously different. The attack capability of adversarial noises comes from the controllable artificial perturbations specified by target DNNs. While the rain effect on DNNs not only depends on the object occlusion and degraded visibility, but also is related to the rain appearances, such as rain intensity, rain direction, and rain brightness, \etc. Which type of rain appearance mostly affects the DNNs, and how the rain appearances cause the failure of DNNs? These are important problems that have so far been relatively little explored, and directly using natural rain to study their effects on different DNN based perception systems is also labor-consuming and expense-costly.

\begin{figure*}[t]
	\centering
  	\includegraphics[width=0.9\linewidth]{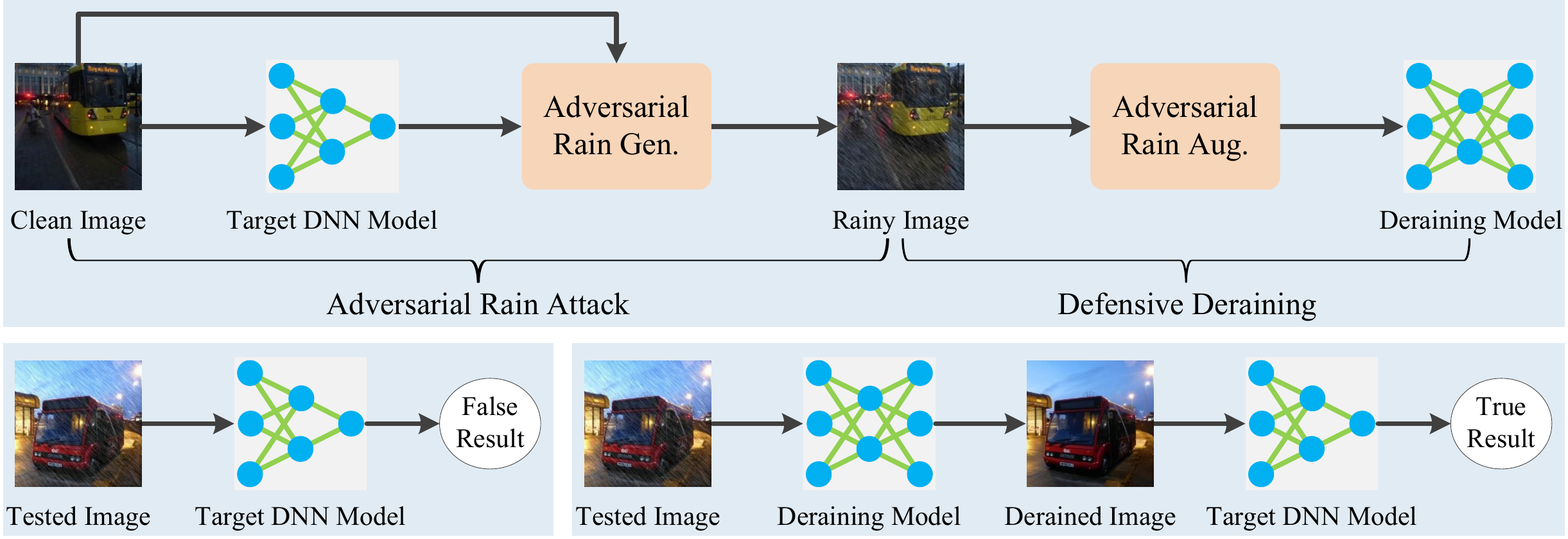}
	\caption{Framework of our methodology. Top: Adversarial rain attack and defensive deraining. Bottom left: DNN system directly processes rainy images and leads to false result. Bottom right: DNN system with defensive deraining provides true result.}
	\label{fig:framework}
\end{figure*}

\begin{figure}[t]
	\centering
  	\includegraphics[width=0.95\columnwidth]{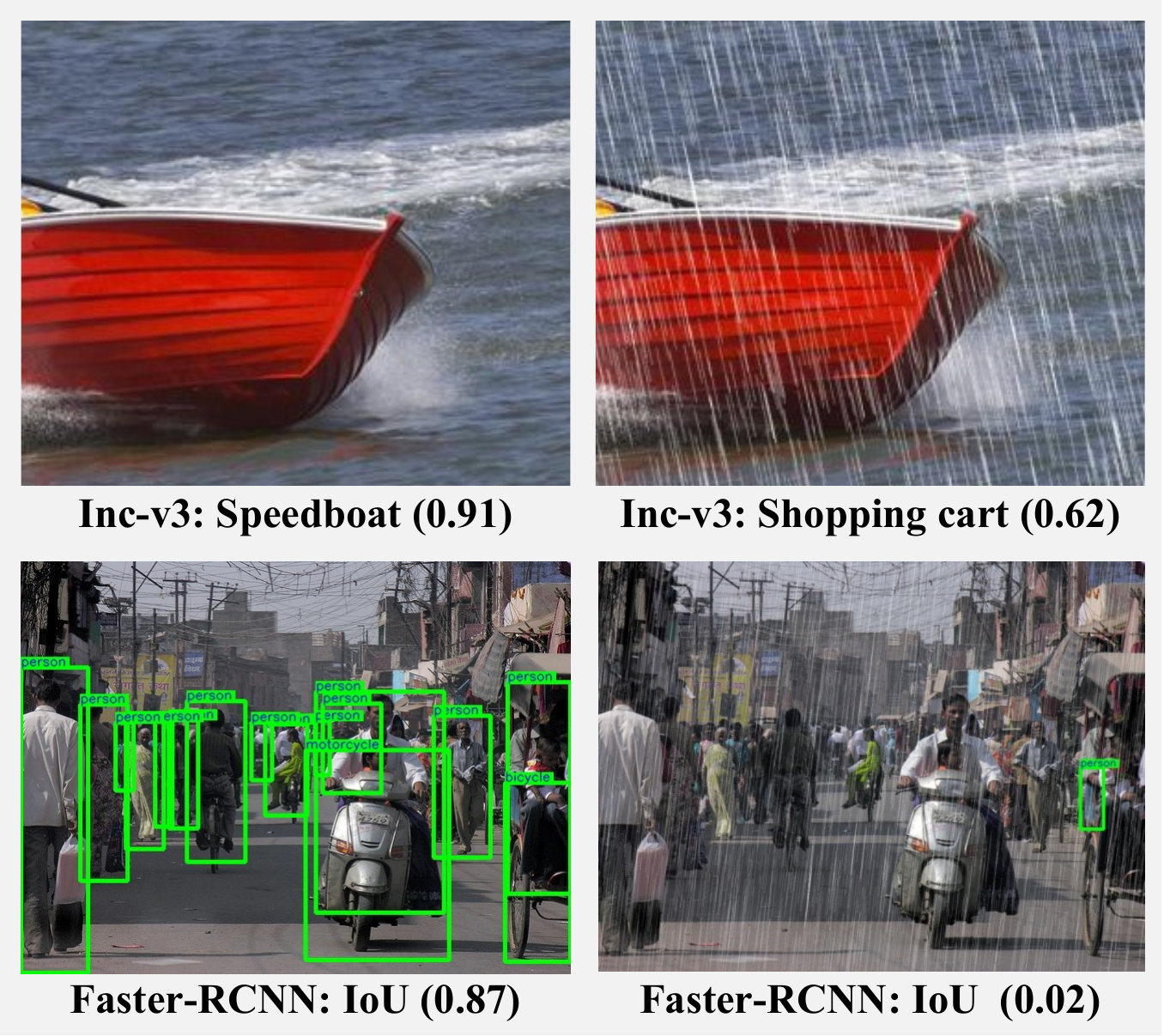}
	\caption{Illustration of clean images (left) and adversarial rain examples (right). First row: images from NeurIPS'17 DEV dataset with image classifier Inception v3 \cite{szegedy2016rethinking}. Second row: images from MS COCO dataset with object detector Faster-RCNN \cite{ren2015faster}.}
	\label{fig:Fig1}
\end{figure}

A common counter-measure to the rain effect on DNNs is the deraining technique \cite{yang2020single}, which removes rain streaks from captured rainy images to restore rain-free images. The deraining task requires large quantities of rainy/clean image pairs for training deraining models, and the rainy images are often artificially synthesized \cite{yang2017deep, zhang2019image, zhang2018density} due to the unavailability of rainy and clean images for the same scene. However, the real rain has various appearances determined by weather conditions and environment lighting, and further undergoes visual changes presented in rainy images owning to varying camera parameters \cite{garg2007vision}. While the synthesized rain emphasizes the rain accumulation that well evaluates the removal capability of deraining models and often neglects the realness and diversity of rain, resulting in a large gap between synthetic rainy scenes and real-word rainy scenes. Even existing rain rendering methods can to some extent generate more natural rainy images, they are mostly used for a specific scenario \cite{garg2006photorealistic,creus2013r4,weber2015multiscale,halder2019physics}. Therefore, it is challenging or even impossible to collect or synthesize the rainy images in all kinds of rain situations that can potentially occur in the real world, and especially the hard rain examples that easily mislead the DNNs.

In this paper, we intend to tackle the above problems altogether from a new angle, by taking a combined perspective of two different studies, \ie, rainy image synthesis and adversarial attack, and present an adversarial rain attack and a defensive deraining strategy, by which we could simulate various (potentially worst-case) rain situations with the guidance of deployed DNNs to reveal the potential threat factors of rain, and in turn we could also train more powerful deraining models with the adversarial rain to defend the deployed DNNs from negative rain effect. The framework of our methodology is shown in Fig. \ref{fig:framework}.
For the adversarial rain attack, we take both adversarial capability and rain appearance quality into consideration. The adversarial rain is made intentionally visible, which is completely opposite to the imperceptible adversarial noises, thus the adversarial ability has to be achieved with stronger constraints.
For this purpose, we design a factor-aware rain generation method that simulates rain streaks following the camera exposure process.
In addition, we also model several types of controllable rain factors, including rain intensity, rain direction and rain brightness, all of which can be learned and tuned for adversarial attack under the constraints of rain appearance.
For the defensive deraining, we propose an adversarial rain augmentation method, which mixes multiple adversarial rain layers to randomize and diversify the rain cases, enhancing the performance of deraining models. Finally, the rain-augmented deraining models can be used as a preprocessor for follow-up perception systems.

We conduct a large-scale evaluation of our adversarial rain attack and defensive deraining. We find that, indeed, our approaches can successfully generate adversarial rainy images with high-quality (see some examples in Fig. \ref{fig:Fig1}), and can benefit the training of deraining models.
Furthermore, our learned rain factors can also be used to guide the generation of real adversarial rain examples, demonstrating the potential threats of natural rain that should be called attention. Our main contributions are summarized as follows:

\begin{itemize}
\item We identify the essential problem of the inevitable rain factors to the impacts and risks of DNNs. With a comprehensive study on the relation of rain effect and DNNs, we propose a new type of adversarial attack by using common rainy weather rather than adversarial noises. The proposed adversarial rain enriches the current family of adversarial examples, in providing an important family of physical effects that can occur in the real world.

\item We design a novel rain generation method that can synthesize photo-realistic rainy images directly from random noises. The synthesized rain exhibits strong DNN attack capability against image classification and object detection, helping to analyze the risks of rain effects to DNNs.

\item We present a defensive deraining strategy, in which the synthesized rain generated from adversarial attack can augment the rainy image datasets, diversifying the rain situations and thus improving the deraining performance for downstream vision tasks.

\item Furthermore, we also propose a simulation approach to launch a real-world adversarial rain attack, from which the real adversarial rain is obtained by the rain factors guided by DNNs, building a bridge between digital adversarial rain and real adversarial rain.


\end{itemize}

The rest of this paper is organized as follows.
In Section II, we summarize the related works on adversarial examples, rain rendering and rain removal methods.
In Section III, we elaborate on the design of our proposed adversarial rain attack and defensive deraining.
In Section IV, we perform extensive experiments to verify the effectiveness of our attack and defensive methods.
Finally, we conclude with some suggestions for future works in Section V.

\begin{figure*}[t]
  \centering
  \includegraphics[width=1.99\columnwidth]{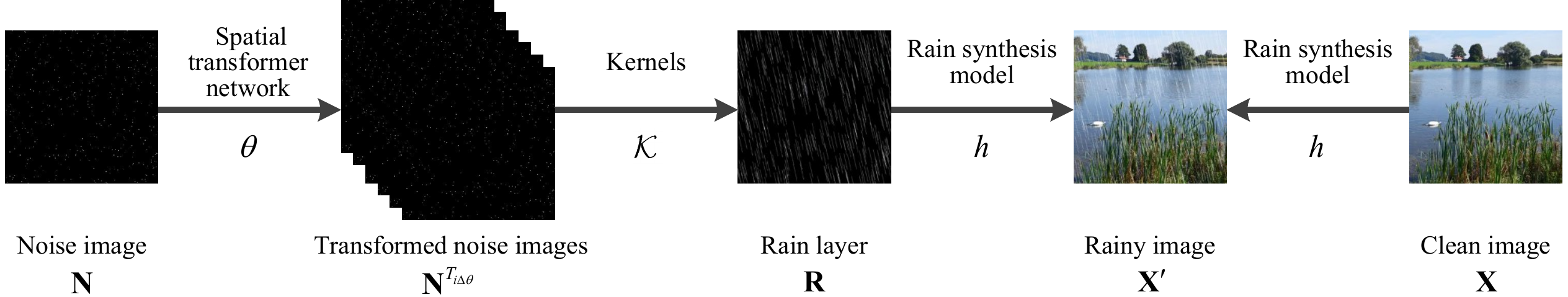}
  \caption{Overview of factor-aware rain generation. Please zoom in for noise and rain details.}
  \label{fig:Fig2}
\end{figure*}

\section{Related Work}\label{sec:related}

\subsection{Adversarial Noise Attacks}
Extensive studies have shown that state-of-the-art DNNs are still vulnerable to adversarial examples \cite{yuan2019adversarial}. In their early study, \cite{szegedy2014intriguing} demonstrated that intentionally designed perturbations added to images can mislead classification results, and proposed to generate adversarial examples by using a box-constrained L-BFGS method. 
To overcome the high complexity in \cite{szegedy2014intriguing}, \cite{goodfellow2015explaining} designed a fast gradient sign method (FGSM) to produce adversarial noises with the sign of loss gradient in a one-step process. Since then, numerous gradient-based adversarial noise attacks for image classification have been proposed, including the attacks by developing the FGSM in multiple iterations \cite{kurakin2016adversarial, carlini2017towards, madry2018towards}, integrating momentum iteratively into FGSM (MI-FGSM) \cite{dong2018boosting}, employing diverse input method (DIM) in each iteration \cite{xie2019improving}, optimizing perturbations with a translation-invariant method (TIM) \cite{dong2019evading}, and incorporating variance tuning into iterative gradient \cite{wang2021enhancing}.

Recently, adversarial examples are also applied to other computer vision tasks. \cite{xie2017adversarial} proposed a dense adversary generation (DAG) algorithm to attack object detection and semantic segmentation models, in which the loss of object proposals with assigned false labels was minimized during gradient-descent iterations. \cite{li2018robust} suggested attacking the region proposal network (RPN) widely adopted in object detectors, and designed a robust adversarial perturbation (RAP) method. \cite{wang2020adversarial} generated adversarial examples by using the total loss of both object detector and RPN. Differently, \cite{wei2019transferable} proposed a unified and efficient adversary (UEA) method for object detection based on a generative adversarial network (GAN), and achieved high transferability and low computation cost. Moreover, recent studies also explore physical adversarial attacks against real-world object detectors \cite{chen2018shapeshifter, huang2020universal}.

The current adversarial attacks are mainly performed under the disguise of additive noises, which are artificial fabrications and may arouse suspicion. We instead use common rainy weather as a natural camouflage for launching adversarial attacks, from which we investigate the relation between attack capability and rain factors.

\subsection{Rain Rendering}
Rain rendering refers to the generation or synthesis of photo-realistic rain effects in computer graphics. Existing rain rendering methods can roughly fall into two categories. The \emph{first category} simulates the texture patterns of rain streaks, and post-processes the rain streaks to fit the scene \cite{wang2004rendering, wang2006real, tatarchuk2006artist}. Such methods are often simple and computationally efficient, but lack of realistic rain appearance. 
The \emph{second category} leverages particle systems, in which the rain rate, rain distribution, and rain velocity can be flexibly controlled \cite{garg2006photorealistic, rousseau2006realistic, creus2013r4, weber2015multiscale, halder2019physics}. In particular, these methods consider the physical and optical properties of rain, being able to generate visually realistic rain effects, but at the expense of high computation costs.

Different from the above texture-based and physics-based rain rendering methods, we generate the rain directly from random noises with the guidance of DNNs, ensuring both high rain realism and low computation complexity. Our noise-generated rain is also completely different from the synthesized rain based on noises used in deraining tasks \cite{luo2015removing, fu2017removing, zhang2018density}, which use a motion-blur kernel to stretch the dotted noises to long rain streaks in one step \cite{ps_rain}. In contrast, we use an $N$-step translation transformation to  transform the noises into rain streaks inspired by the camera exposure process, enabling flexible and controllable rain generation with learnable network parameters and also effective adversarial attack capability.

\subsection{Rain Removal}
Single image rain removal is the opposite of rain rendering, and can also be largely divided into two categories \cite{yang2020single}. The \emph{first category} is model-based methods, regarding the single image rain removal as an image decomposition or signal separation problem between the rain layer and background layer. Typical methods include dictionary-based sparse coding \cite{kang2011automatic, luo2015removing}, nonlocal mean filtering \cite{kim2013single}, low rank representation model \cite{chen2013generalized, chang2017transformed}, and patch-based Gaussian mixture model \cite{li2016rain}.

The \emph{second category} is the recently fast-growing deep learning-based methods, which automatically learn a non-linear mapping between rainy images and clean images. Many DNN architectures have been proposed for rain removal.
Some representative methods include deep detail network for predicting rain residues \cite{fu2017removing}, multi-stream dense network for rain density estimation \cite{zhang2018density}, multi-task network for joint rain detection and removal \cite{yang2017deep}, scale-free network for two-stage deraining \cite{yang2019scale}, progressive recurrent network for multi-stage rain removal \cite{ren2019progressive}, and pixel-wise filter learning for efficient deraining \cite{guo2020efficientderain}. Furthermore, some recent attempts are also made by using GANs~\cite{zhang2019image, qian2018attentive, li2019heavy} and semi-supervised learning~\cite{ wei2019semi, yasarla2020syn2real} for the deraining task.


The existing deraining models are trained on synthetic rainy images, which cannot contain all kinds of rain cases. In the experimental section, we use our adversarial rain to augment the rainy image datasets during the training to improve the deraining performance.

\section{Methodology}\label{sec:method}
In this section, we first introduce the factor-aware rain generation, which synthesizes the rainy images to be prepared for adversarial attack. Next, we detail the implementation of adversarial rain attack on image classification and object detection tasks. Finally, we describe the defensive deraining strategy equipped with the adversarial rain augmentation.

\subsection{Factor-aware Rain Generation}

Natural rain is liquid water that falls visibly in small drops. Due to the high speeds of drops and the exposure time limitation of a camera, the rain in an image can usually be motion-blurred and displayed as longer streaks \cite{garg2007vision}.
The latest rain rendering methods \cite{halder2019physics,creus2013r4} are mostly based on a ready-made rain streak database \cite{garg2006photorealistic} and/or a particle simulator, requiring strict operating conditions. We instead propose a new mechanism, named factor-aware rain generation, to generate photo-realistic rain streaks (see Fig. \ref{fig:Fig2}). We directly leverage random noises as a starting point to generate rain streaks, by taking the following consideration.

\begin{itemize}
\item A raindrop typically has a diameter of 1-2 mm, and usually does not exceed 10 mm \cite{garg2007vision, van1997numerical}. The tiny raindrops in a far distance captured by the camera look like image noises. Some deraining methods also treat the raindrops or rain streaks in images as special high-frequency noises \cite{chang2017transformed, eigen2013restoring}. These are important physical rain properties and should be considered during method design for better photo-realisticity.
\item Many synthesized rain databases commonly adopt random noises as the basis \cite{fu2017removing,zhang2018density}, which can be an effective starting point.
\item Random noises can be easily modulated during gradient back-propagation, which provides the chance and efficient way for the combination investigation of worst-case rain effects to the decision of DNNs.
\end{itemize}

To generate realistic-looking rain from random noises, some properties of noises, including noise distribution, noise strength and noise density, should be carefully considered. The random noises are generated from a uniform distribution ${\cal U}(a,b)$ and form a noise image with the same size as the input clean image. Since the rain streak usually has larger brightness intensities, the lower bound $a$ and upper bound $b$ of ${\cal U}(a,b)$ are assigned large values (see experimental section). The noise image is then uniformly sampled using a sampling rate $\varepsilon _n$ by setting the non-selected elements to zeros, thus obtaining a sparse noise image. The sampling rate $\varepsilon _n$ is the ratio of the desired number of noise elements to the image size, and it determines the noise density of a noise image, and also affects the rain rate in the final rainy image. The reason behind the uniform distribution and uniform sampling is that the distribution of drops is uniform over space and time \cite{garg2007vision}.

Inspired by the formation of rain streaks in camera sensing process, we simulate the motion-blur of the noise image as the rain effects. Since the motion-blur is an integration of the positions of moving objects over the period of exposure time, the motion-blur of the noise image can be simulated by continuous translation along a specific direction.

Let $\mathbf N$ be a noise image, and $T_ \theta$ the translation transformation with translation parameters $\theta$. The transformed noise image is denoted as ${\mathbf N}^{T_\theta }$. The $T_ \theta$ can be performed by using a spatial transformer network \cite{jaderberg2015spatial}, and thus $\theta$ represents affine matrix parameters. To transform the noises into continuous rain streaks, we further divide the translation transformation process into $N$ parts. To this end, we apply an $N$-step translation transformation to $\mathbf N$ using translation parameters $i\Delta \theta$, where $\Delta \theta = \theta / N$ and $i \in \left[ {0,N} \right]$. 
After the $N$-step translation transformation, we obtain $N$ transformed noise images ${\mathbf N}^{{T_{i\Delta \theta }}}$. To smooth the rain streak formed by summing up $N$ transformed noise images, we convolve the $N$ transformed noise images with a kernel ${\mathbf K}_p$ with size $(N \!+\! 1) \!\times\! C \!\times\! 1 \!\times\! 1$, where $C$ is the number of color channels. Finally, the $p$-th rain streak ${\mathbf R}_p$ is generated by
\begin{equation}
  {{\mathbf R}_p} = g({{\mathbf N}_p},{T_\theta },{{\mathbf K}_p}) = \sum\limits_{i,q \in [0,N]} {{\mathbf N}_p^{{T_{i\Delta \theta }}}{k_{pq}}}
  \label{eq:rain_gen}
\end{equation}
where ${\mathbf N}_p$ is the $p$-th noise element in $\mathbf N$, ${\mathbf K}_p$ is the kernel corresponding to ${\mathbf N}_p$, and ${k_{pq}}$ is the $q$-th weight of ${\mathbf K}_p$. All the ${\mathbf R}_p$ make up a rain layer $\mathbf R$.

Given a clean image $\mathbf X$ and a rain layer $\mathbf R$, the rainy image $\mathbf X'$ can be synthesized by various rain synthesis models. Take the additive composite model \cite{yang2020single} as an example, the rainy image synthesis is formulated as
\begin{equation}
  {\mathbf X'} = h({\mathbf X}, {\mathbf R}) = {\mathbf X} + {\mathbf R}
  \label{eq:rain_model}
\end{equation}
where $h(\cdot, \cdot)$ represents a synthesis function.

Note that, in this initial attempt, we have not yet considered environmental lighting for rain generation, and we use a simple scenario where the rain streaks are thin enough to neglect the reflection and refraction of light. 
More complicated and advanced can be integrated into our rain generation process \cite{garg2007vision,halder2019physics} as future work.

The factor-aware rain generation method just provides initial rainy images, which still lack of adversarial ability. In the following two subsections, we use the DNNs deployed in image classification and object detection to further guide the rain factors for producing adversarial rainy images.

\begin{algorithm}[t]
\footnotesize
  \caption{Adversarial Rain Attack on Image Classification}
  \label{alg:alg_1}
  \KwIn{Clean image $\mathbf X$, classifier $f$ with loss function $L$, rain factors $\mathbf N$, $\theta$ and ${\cal K}$}
  \KwOut{Adversarial rainy image $\mathbf X'$}
  
  Initialize ${{\mathbf N}_{(1)}} \!=\! {\mathbf N}$, ${{\theta}_{(1)}} \!=\! {\theta}$, ${{\cal K}_{(1)}} \!=\! {\cal K}$, ${L_{max}} \!=\! 0$\;
  
  \For{$s = 1, \cdots ,S$}{
  
  Generate a rain layer ${\mathbf R}_{(s)}$ using ${\mathbf N}_{(s)}$, ${\theta}_{(s)}$ and ${\cal K}_{(s)}$ based on Eq. (\ref{eq:rain_gen})\;
  
  Synthesize a rainy image ${\mathbf X'}_{(s)}$ using $\mathbf X$ and ${\mathbf R}_{(s)}$ based on Eq. (\ref{eq:rain_model})\;
  
  Get the loss $L({\mathbf X'}_{(s)},y)$ from $f({\mathbf X'}_{(s)})$, and calculate the gradient ${\nabla _{{\mathbf X'}_{(s)}}}L({\mathbf X'}_{(s)},y)$ of $L({\mathbf X'}_{(s)},y)$ w.r.t. ${\mathbf X'}_{(s)}$\;
  
  Back-propagate the sign of gradient ${\nabla _{{\mathbf X'}_{(s)}}}L({\mathbf X'}_{(s)},y)$ to the rain factors ${\mathbf N}_{(s)}$, ${\theta}_{(s)}$ and ${\cal K}_{(s)}$, and update them as ${\mathbf N}_{(s+1)}$, ${\theta}_{(s+1)}$ and ${\cal K}_{(s+1)}$ via a gradient ascent method\;
  
  \If{$L({\mathbf X'}_{(s)},y) > {L_{max}}$}{
  ${L_{max}} = L({\mathbf X'}_{(s)},y)$\;
  ${{\mathbf X'}_{opt}} = {{\mathbf X'}_{(s)}}$\;
  }
  }
  \textbf{return} ${{\mathbf X}'} = {{\mathbf X'}_{opt}}$\;
\end{algorithm}

\subsection{Adversarial Rain Attack on Classification}
Given a classifier $f({\mathbf X}): {\mathbf X} \in {\cal X} \to y \in {\cal Y}$ that maps an input image $\mathbf X$ to a ground-truth class label $y$. The adversarial attack aims to generate a constrained example $\mathbf X'$ that can mislead $f({\mathbf X'})$ to output a false label, usually by maximizing the loss $L({\mathbf X'},y)$.

In conventional adversarial noise attacks, the adversarial example $\mathbf X'$ is typically generated by adding adversarial noises to $\mathbf X$. For our adversarial rain attack, we generate $\mathbf X'$ by using the rain generation (Eq.~\eqref{eq:rain_gen}) and rain synthesis model (Eq.~\eqref{eq:rain_model}). To enable successful adversarial attacks of rainy images for image classification, we need to optimize the rain factors under the constraint of rain appearance. There are three types of rain factors should be considered: noise $\mathbf N$, translation parameters $\theta$ and kernels ${\cal K} = \left\{ {{{\mathbf K}_p}|p \in \left[ {1,{{\left\| {\mathbf N} \right\|}_0}} \right]} \right\}$. The $\mathbf N$ affects the rain intensity, $\theta$ controls the direction of rain streaks, and $\cal K$ determines the brightness and smoothness of rain streaks. The adversarial rainy images can be generated by solving the following constrained optimization problem:
\begin{equation}
  \begin{aligned}
  & \! \mathop {\argmax }\limits_{{\mathbf N}, \, \theta, \, {\cal K}} L(h({{\mathbf X}_p}, \sum\limits_{i,q \in [0,N]} {{\mathbf N}_p^{{T_{i\Delta \theta }}}{k_{pq}}} ),\;y) \\
  & {\rm{s.t.}}\quad \forall p,\;{\left\| {\mathbf N} \right\|_0} \le {\varepsilon _n}M,\; \theta \le {\varepsilon _\theta }, \; {k_{pq}} \le {\varepsilon _k}
  \end{aligned}
  \label{eq:loss_func_IC}
\end{equation}
where $M$ is the number of the pixels in a clean image, $\varepsilon _n$ is the sampling rate mentioned earlier, and $\varepsilon _\theta$ and $\varepsilon _k$ denote the thresholds for $\theta$ and $k_{pq}$. For natural rain, the rain streaks often exhibit certain ranges of intensity (relevant to rain rate), directions (affected by wind) and brightness (depending on environment lighting and camera settings). The $\varepsilon _n$, $\varepsilon _\theta$ and $\varepsilon _k$ are used to curb the rain factors to ensure realistic looking rain, and their values will be discussed in the experimental section.

During the rain generation process, we can calculate the gradient of the loss function \wrt all rain factors, and tune the rain factors along a gradient ascent direction for adversarial attacking. Therefore, the adversarial rain attack can be integrated into current gradient based adversarial noise attack frameworks, such as FGSM \cite{goodfellow2015explaining}, MI-FGSM \cite{dong2018boosting}, DIM \cite{xie2019improving} and TIM \cite{dong2019evading}, \etc.

Algorithm \ref{alg:alg_1} presents the details of our adversarial rain attack for the image classification task. In particular, each iteration generates a loss, a rain layer and a rainy image. Eventually, only the rainy image with the largest loss in $S$ iterations is selected as the final adversarial rainy image.

\begin{algorithm}[t]
\footnotesize
  \caption{Adversarial Rain Attack on Object Detection}
  \label{alg:alg_2}
  \KwIn{Clean image $\mathbf X$, detoctor $f$ with loss function $L_j$ ($j=1, \cdots ,J$), rain factors $\mathbf N$, $\theta$ and ${\cal K}$}
  \KwOut{Adversarial rainy image $\mathbf X'$}
  
  Initialize ${{\mathbf N}_{(1)}} \!=\! {\mathbf N}$, ${{\theta}_{(1)}} \!=\! {\theta}$, ${{\cal K}_{(1)}} \!=\! {\cal K}$, ${L_{sum}} \!=\! 0$, ${L_{min}} \!=\! {+ \infty}$\;
  
  \For{$s = 1, \cdots ,S$}{
  
  Generate a rain layer ${\mathbf R}_{(s)}$ using ${\mathbf N}_{(s)}$, ${\theta}_{(s)}$ and ${\cal K}_{(s)}$ based on Eq. (\ref{eq:rain_gen})\;
  
  Synthesize a rainy image ${\mathbf X'}_{(s)}$ using $\mathbf X$ and ${\mathbf R}_{(s)}$ based on Eq. (\ref{eq:rain_model})\;
  
  \For{$j = 1, \cdots ,J$}{
  
  Get the loss $L_j({\mathbf X'}_{(s)},y)$ from $f({\mathbf X'}_{(s)})$, and calculate the gradient ${\nabla _{{\mathbf X'}_{(s)}}}L_j({\mathbf X'}_{(s)},y)$ of $L_j({\mathbf X'}_{(s)},y)$ w.r.t. ${\mathbf X'}_{(s)}$\;
  
  Back-propagate the sign of gradient ${\nabla _{{\mathbf X'}_{(s)}}}L_j({\mathbf X'}_{(s)},y)$ to the rain factors ${\mathbf N}_{(s)}$, ${\theta}_{(s)}$ and ${\cal K}_{(s)}$, and update them as ${\mathbf N}_{(s+1)}$, ${\theta}_{(s+1)}$ and ${\cal K}_{(s+1)}$ using a gradient descent method\;
  
  $L_{sum} = L_{sum} + L_j({\mathbf X'}_{(s)},y)$
  }
  
  \If{$L_{sum} < {L_{min}}$}{
  ${L_{min}} = L_{sum}$\;
  ${{\mathbf X'}_{opt}} = {{\mathbf X'}_{(s)}}$\;
  }
  }
  \textbf{return} ${{\mathbf X}'} = {{\mathbf X'}_{opt}}$\;
\end{algorithm}

\subsection{Adversarial Rain Attack on Object Detection}
In this subsection, we further extend our adversarial rain attack to the object detection task. Given an object detector $f({\mathbf X}):{\mathbf X}\in {\cal X} \to t \in {\cal T}$ that maps an input image $\mathbf X$ to a recognition target $t$. The $t$ can be of any intermediate outputs or final outputs of a detector, not limited to the class labels \cite{xie2017adversarial}, such as final bounding-boxes \cite{wang2020adversarial}, confidence scores and the bounding-boxes of the object proposals in an region proposal network (RPN) widely adopted in detectors \cite{li2018robust}. All these types of targets can be utilized to generate constrained adversarial examples $\mathbf X'$ to fool $f({\mathbf X'})$. This process can be accomplished by minimizing the loss $L({\mathbf X'},{\cal T}')$, where ${\cal T}'$ is a set of one type of adversarial targets $t'$.

Similar to the image classification task, the adversarial rainy images for object detection can be generated by
\begin{equation}
  \begin{aligned}
  & \! \mathop {\argmin }\limits_{{\mathbf N}, \, \theta, \, {\cal K}} \sum\limits_j {{L_j}(h({{\mathbf X}_p},\sum\limits_{i,q \in [0,N]} {{\mathbf N}_p^{{T_{i\Delta \theta }}}{k_{pq}}} ),\;{{{\cal T}_j}'})} \\
  & {\rm{s}}{\rm{.t}}{\rm{.}}\quad \forall p,\;{\left\| {\mathbf N} \right\|_0} \le {\varepsilon _n}M,\;\theta  \le {\varepsilon _\theta },\;{k_{pq}} \le {\varepsilon _k}
  \end{aligned}
  \label{eq:loss_func_OD}
\end{equation}
where $j$ is the index of adversarial target types (\eg, class labels, bounding-boxes, or confidence scores). One or more types of adversarial target set ${{\cal T}_j}'$ can be considered for the total loss function, in which $L_j$ denotes the loss function corresponding to ${{\cal T}_j}'$.

Similarly, we can tune the rain factors via a gradient descent algorithm, and also apply the adversarial rain attack to current gradient based adversarial noise attacks for object detection, such as DAG \cite{xie2017adversarial} and RAP \cite{li2018robust}, \etc.
The adversarial rain attack on object detection is summarized in Algorithm \ref{alg:alg_2}, and the attacking process is similar to that of image classification.

\begin{algorithm}[t]
\footnotesize
  \caption{Adversarial Rain Augmentation}
  \label{alg:alg_3}
  \KwIn{Clean image $\mathbf X$, rainy image ${\mathbf X}^{\rm{r}}$}
  \KwOut{Augmented rainy image ${\mathbf X}^{\rm{ar}}$}
  
  Initialize an empty rain layer ${\mathbf R}$ with all zeros;
  
  Sample mixing weights $\left( {{w_1},{w_2}, \cdots ,{w_k}} \right) \sim {\rm{Dirichlet}}$\;
  
  \For{$i = 1, \cdots ,k$}{
  
  Generate an Adversarial rain layer ${\mathbf R}_{i}$ with new noise initialization\;
  
  ${\mathbf R} = {\mathbf R} + {w_i}{\mathbf R}_{i}$\;
  }
  
  
  Sample ${\hat {\mathbf X}} \sim \left\{{\mathbf X}, {\mathbf X}^{\rm{r}} \right\}$\;
  
  
  \textbf{return} ${\mathbf X}^{\rm{ar}} = {\hat {\mathbf X}} + {\mathbf R}$\;
\end{algorithm}

\subsection{Defensive Deraining}
Rain causes severe visual degradation to captured images and thus poses a risk for follow-up perception tasks, so rain removal is quite necessary for DNN based perception systems in practical applications. However, existing deraining models are always trained on synthesized rainy images in limited situations, which cannot exhaustively represent all the possible rain patterns (especially the hard rain examples for DNNs), thus limiting the deraining performance.


In previous subsections, we have elaborated on the adversarial rain attack for different vision tasks, and this motivates us to use it as a defense to improve the performance of DNN perception. Our defensive deraining strategy consists of two stages: we first leverage the adversarial rainy images to augment the rain dataset to train more powerful deraining models, and then we use the trained deraining model to remove the rain streaks in the rainy images which are inputs for DNNs. The challenge of the defensive deraining strategy is to augment the rain dataset with enough diverse rain situations, which should encourage the deraining models to focus more on the hard-to-remove rain streaks that are more important for perception recognition.

To address the challenge and to close the gap between synthesized rain and real rain for training de-raining models, we propose an adversarial rain augmentation method, which is inspired by AugMix \cite{hendrycks2019augmix}. Specifically, for each rainy/clean image pair during the training, we generate $k$ different adversarial rain layers for the clean image to diversify the rain cases. Then we aggregate the adversarial rain layers with the weights sampled from a Dirichlet distribution for a random blending. Finally we obtain the augmented rainy image by adding the aggregated adversarial rain layer to the clean image or rainy image.
Different from AugMix \cite{hendrycks2019augmix} that mixes basic transformations to improve model robustness, our adversarial rain augmentation uses mixed adversarial rain layers, aiming to diversify the rain to match the real rain situations. The detailed data augmentation process is described in Algorithm \ref{alg:alg_3}. Our augmentation method can be applied to existing rain datasets, and see the experiments in Section \ref{exp_deraining}.

\begin{figure}[t]
	\centering
  	\includegraphics[width=0.95\columnwidth]{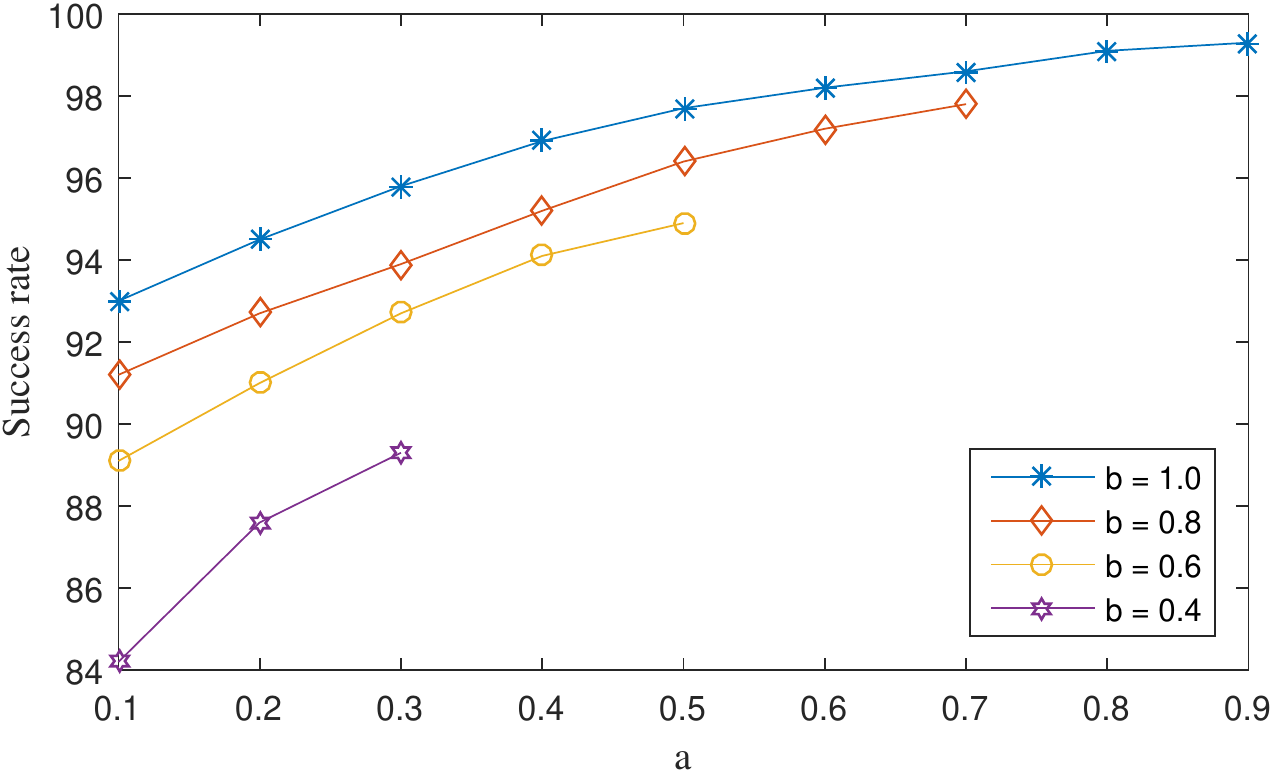}
	\caption{The effect of ${\cal U}(a,b)$ on adversarial capability.}
	\label{fig:uniform_para}
\end{figure}

\section{Experiments}\label{sec:exp}

\subsection{Experimental Setup}
\subsubsection{Datasets}
We perform a comprehensive evaluation of the adversarial rain attack on two different tasks: image classification and object detection. For image classification, we use NeurIPS'17 adversarial competition dataset DEV \cite{kurakin2018adversarial} for experiments, which consists of 1,000 images, and is compatible with ImageNet. For object detection, we perform experiments on MS COCO 2014 minival split \cite{lin2014microsoft} and KITTI object benchmark \cite{geiger2012we}, which contain 5,000 images and 7,480 images, respectively. We also use rain datasets Rain100H \cite{yang2017deep}, Rain800 \cite{zhang2019image}, Rain1200 \cite{zhang2018density} and Rain1400 \cite{fu2017removing} for rain appearance comparison and deraining experiments.


\begin{table}[t]
  \centering
    \begin{minipage}{\columnwidth}
    \newcommand{\tabincell}[2]{\begin{tabular}{@{}#1@{}}#2\end{tabular}}
    \centering
    \caption{The effect of $\varepsilon_n$, $\varepsilon_\theta$, and $\varepsilon_k$ on adversarial capability.}
    \label{tab:para_threshold}
    \resizebox{0.95\linewidth}{!}{
    \begin{tabular}{c|cccccc}
    \toprule
    $\varepsilon _n$ & 0.001 & 0.0025 & 0.005 & 0.0075 & 0.01 & 0.02 \\
    Success rate & 81.5  &   90.3 & 98.6  &   99.4 & 99.8 & 99.9 \\
    \midrule
    $\varepsilon _\theta$ & 0.01 & 0.05 &  0.1 &  0.2 &  0.3 & 0.4 \\
    Success rate          & 85.4 & 92.2 & 96.0 & 98.6 & 99.2 & 9.5 \\
    \midrule
    $\varepsilon _k$ & 0.05 &  0.1 &  0.2 &  0.3 &  0.4 &  0.5 \\
    Success rate     & 82.3 & 89.8 & 94.7 & 98.6 & 99.7 & 99.9 \\ 
    \bottomrule
    \end{tabular}
    }
  \end{minipage}
  \hfill
  \vspace{10pt}
  \begin{minipage}{\columnwidth}
    \newcommand{\tabincell}[2]{\begin{tabular}{@{}#1@{}}#2\end{tabular}}
    \caption{The success rates (in \%) of normal rain and AdvRain under the same rain magnitude on NeurIPS'17 DEV.}
    \label{tab:rain_mag_DEV}
    \centering
    \resizebox{0.95\linewidth}{!}{
    \begin{tabular}{lcccc}
    \toprule
    \multirow{2}{*}[-2pt]{\tabincell{c}{Rain\\Methods}} & \multicolumn{4}{c}{Threat Models (Inc-v3)} \\
    \cmidrule(l){2-5}
    & Inc-v3 & Inc-v4 & IncRes-v2 & Xception \\
    \midrule
    NormalRain1 & 39.3 & 38.5 & 36.4 & 38.8 \\
    AdvRain1    & \textbf{81.5} & \textbf{46.7} & \textbf{41.8} & \textbf{53.0} \\
    \midrule
    NormalRain2 & 45.5 & 43.2 & 41.6 & 44.0 \\
    AdvRain2    & \textbf{98.6} & \textbf{54.3} & \textbf{50.1} & \textbf{62.5} \\
    \bottomrule
    \end{tabular}
    }
  \end{minipage}
\end{table}

\begin{figure*}[t]
  \centering
  \includegraphics[width=\linewidth]{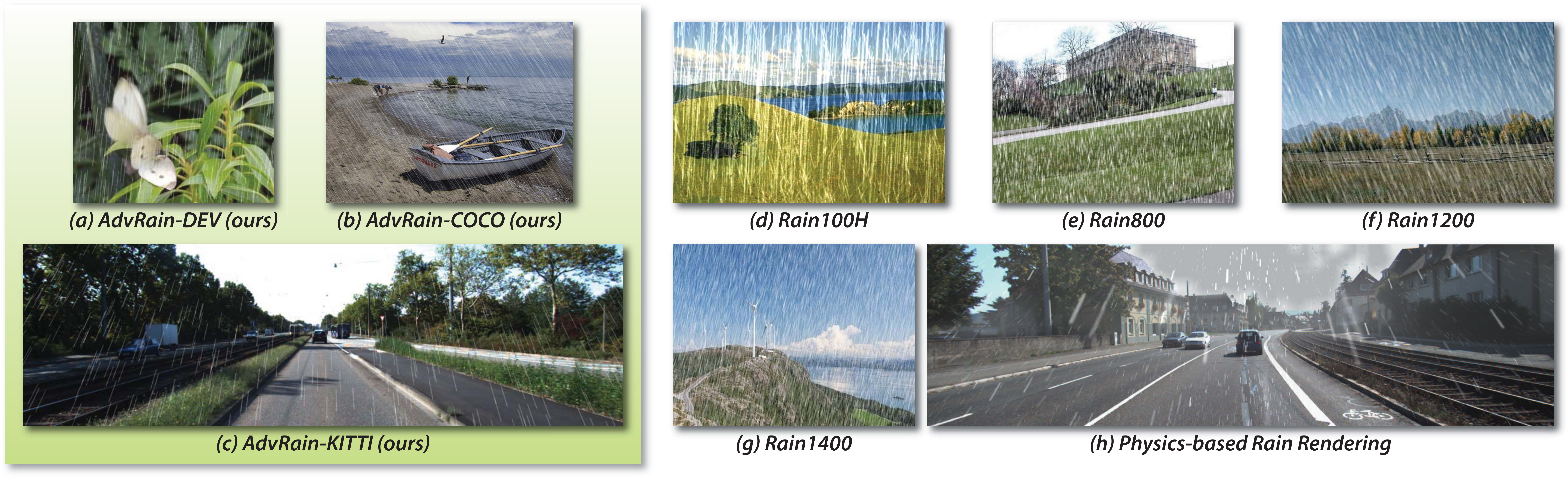}
  \caption{Visual comparison of our adversarial rainy images (a-c) and other synthesized rainy images from Rain100H \protect\cite{yang2017deep}, Rain800 \protect\cite{zhang2019image}, Rain1200 \protect\cite{zhang2018density}, Rain1400 \protect\cite{fu2017removing} and physics-based rain rendering \protect\cite{halder2019physics} (d-h).}
  \label{fig:Eva_figs}
\end{figure*}

\subsubsection{Threat Models}
To evaluate our adversarial rain attack on image classification, we exploit four publicly-available pre-trained models, including Inception v3 (Inc-v3) \cite{szegedy2016rethinking}, Inception v4 (Inc-v4), Inception ResNet v2 (IncRes-v2) \cite{szegedy2017inception}, and Xception \cite{chollet2017xception}. 
The threat model for object detection is Faster RCNN (FR) \cite{ren2015faster} with different backbones, including VGG16 (v16) \cite{simonyan2014very}, MobileNet (mn) \cite{howard2017mobilenets}, ResNet50 (rn50), ResNet101 (rn101) and ResNet152 (rn152) \cite{he2016deep}.

\subsubsection{Baselines}
We compare the adversarial rain attack with six adversarial noise attacks on image classification: FGSM \cite{goodfellow2015explaining}, MI-FGSM \cite{dong2018boosting}, DIM \cite{xie2019improving} and TIM’s three variants TI-FGSM, TI-MI-FGSM, and TI-DIM \cite{dong2019evading}. The comparative methods used for object detection are RAP \cite{li2018robust} and UEA \cite{wei2019transferable}.

\subsubsection{Evaluation Metrics}
We use attack success rate and mean average precision (mAP) as major metrics to measure the adversarial capability on image classification and object detection, respectively.

\subsubsection{Implementation Details}
To attack image classifiers and object detectors, we individually generate adversarial rain examples by integrating with different existing adversarial attacks (MI-FGSM \cite{dong2018boosting}, DIM \cite{xie2019improving} and TI-MI-FGSM \cite{dong2019evading} for classification, and RAP \cite{li2018robust} for detection). All the adversarial rain examples are obtained in $S=20$ iterations, and the other hyper-parameters are consistent with the default settings in \cite{dong2018boosting, xie2019improving, dong2019evading, li2018robust}.

\begin{figure}[t]
	\centering
  	\includegraphics[width=\columnwidth]{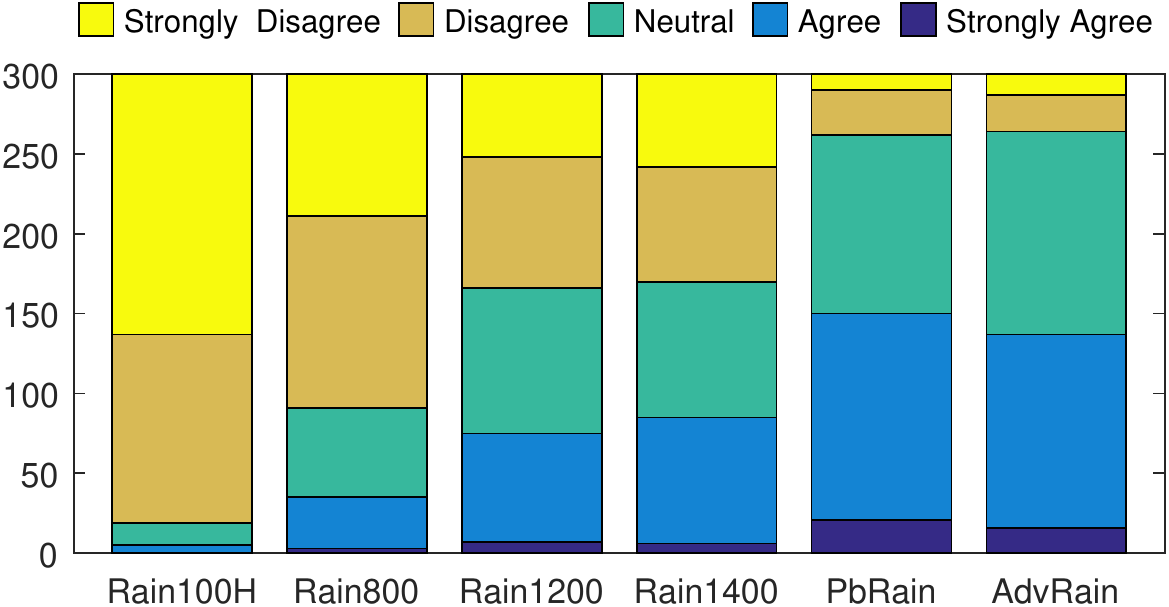}
	\caption{User study of visual realism for different rainy images.}
	\label{fig:user_study}
\end{figure}

\subsection{Ablation Study}
\subsubsection{Rain Factors}
We first conduct ablation experiments to study the impact of rain factors on adversarial rain attack. When considering one rain factor, the other rain factors are set as default values and remain unchanged. For simplicity, all the ablation experiments are only performed on the image classification task.

As described in Section \ref{sec:method}, the adversarial rain is generated from random noises that follow a normal distribution ${\cal U}(a,b)$, and further optimized under the constraints related to thresholds $\varepsilon _n$, $\varepsilon _\theta$ and $\varepsilon _k$. We first show in Fig. \ref{fig:uniform_para} the effects of different combinations of lower bound $a$ and upper bound $b$ in ${\cal U}(a,b)$ on the adversarial capability. We observe that the combinations of $a$ and $b$ with larger values contribute to a higher attack success rate due to larger noise strength. A larger distribution interval increases noise diversity, and thus enriches the texture pattern of rain appearance. Eventually, we choose ${\cal U}(0.7,1.0)$ as a suitable compromise.

The $\varepsilon _n$ and $\varepsilon _\theta$ are scalar thresholds, and the $\varepsilon _k$ is a vector containing thresholds for translation parameters in affine transform. We report the effects of three types of thresholds $\varepsilon _n$, $\varepsilon _\theta$ and $\varepsilon _k$ on the adversarial capability in Table \ref{tab:para_threshold}. We can see that the adversarial rain attack achieves higher attack success rates at larger $\varepsilon _n$, $\varepsilon _\theta$ and $\varepsilon _k$, indicating that higher rain intensity, longer rain streaks and brighter rain textures affect the DNNs more severely. However, the attack success rate saturates at larger threshold values, which also degrades the rain appearance. Therefore, we select $\varepsilon _n = 0.005 $, $\varepsilon _\theta = 0.2$ and $\varepsilon _k = 0.3$ for following experiments.

\subsubsection{Adversarial Rain v.s. Normal Rain} \label{advrain_normalrain}
To validate whether or not the high attack performance of our adversarial rain is gained by the large magnitude of the perturbation, we compare the classification results of normal rain \cite{garg2006photorealistic} and adversarial rain under the same rain magnitude in Table \ref{tab:rain_mag_DEV}. 

Specifically, we consider two rain magnitudes for the normal rain and adversarial rain, and the magnitudes are decided by the area of rain streaks in the binary rain maps. The AdvRain1 and AdvRain2 denote the adversarial rain with noise density $\varepsilon _n = 0.001$ and $\varepsilon _n = 0.005$ respectively, and NormalRain1 and NormalRain2 have the same rain magnitudes corresponding to AdvRain1 and AdvRain2.
We can see from Table \ref{tab:rain_mag_DEV} that the adversarial rain achieves consistently higher attack success rates than the normal rain in all cases, indicating that our superior attack performance does come from adversarially learned rain factors rather than large rain magnitude.

\begin{table*}[t]
  \centering
  \begin{minipage}{\columnwidth}
    \newcommand{\tabincell}[2]{\begin{tabular}{@{}#1@{}}#2\end{tabular}}
    \caption{The success rates (in \%) of various adversarial attack methods against image classifiers on NeurIPS'17 DEV.}
    \label{tab:Adv_IC_DEV}
    \centering
    \resizebox{0.9\linewidth}{!}{
    \begin{tabular}{lcccc}
    \toprule
    \multirow{2}{*}[-2pt]{\tabincell{c}{Attack\\Methods}} &   \multicolumn{4}{c}{Threat Models (Inc-v3)} \\
    \cmidrule(l){2-5}
    & Inc-v3 & Inc-v4 & IncRes-v2 & Xception \\
    \midrule
    FGSM \cite{goodfellow2015explaining}
    & 79.6 & 35.9 & 30.6 & 42.1 \\
    MI-FGSM \cite{dong2018boosting}
    & 97.8 & 47.1 & 46.4 & 47.7 \\
    AdvRain-MI-FGSM
    & \textbf{98.6} & \textbf{54.3} & \textbf{50.1} & \textbf{62.5} \\
    \midrule
    DIM \cite{xie2019improving}
    & 98.3 & 73.8 & 67.8 & 71.6 \\
    AdvRain-DIM
    & \textbf{98.5} & \textbf{76.4} & \textbf{69.5} & \textbf{74.2} \\
    \midrule
    TI-FGSM \cite{dong2019evading}
    & 75.4 & 37.3 & 32.1 & 38.6 \\
    TI-MI-FGSM \cite{dong2019evading}
    & 97.9 & 52.4 & 47.9 & 44.6 \\
    TI-DIM \cite{dong2019evading}
    & 98.5 & 75.2 & 69.2 & 61.3 \\
    AdvRain-TI-MI-FGSM
    & \textbf{98.8} & \textbf{79.8} & \textbf{74.3} & \textbf{66.2} \\
    \bottomrule
    \end{tabular}
    }
  \end{minipage}
  \hfill
  \begin{minipage}{\columnwidth}
    \newcommand{\tabincell}[2]{\begin{tabular}{@{}#1@{}}#2\end{tabular}}
    \caption{The success rates (in \%) of various adversarial attack methods against defense models on NeurIPS'17 DEV.}
    \label{tab:Adv_defense}
    \centering
    \resizebox{0.9\linewidth}{!}{
    \begin{tabular}{lccc}
    \toprule
    \multirow{2}{*}[-2pt]{\tabincell{c}{Attack\\Methods}} & \multicolumn{3}{c}{Threat Models (Inc-v3)} \\
    \cmidrule(l){2-4}
    & Inc-v3$\rm{_{ens3}}$ & Inc-v3$\rm{_{ens4}}$ & IncRes-v2$\rm{_{ens}}$ \\
    \midrule
    FGSM \cite{goodfellow2015explaining}
    & 13.7 & 14.5 & 6.8 \\
    MI-FGSM \cite{dong2018boosting}
    & 15.9 & 16.3 & 7.0 \\
    AdvRain-MI-FGSM 
    & \textbf{65.7} & \textbf{68.2} & \textbf{59.2} \\
    \midrule
    DIM \cite{xie2019improving}
    & 17.9 & 21.8 & 9.7 \\
    AdvRain-DIM 
    & \textbf{67.6} & \textbf{69.5} & \textbf{61.4} \\
    \midrule
    TI-FGSM \cite{dong2019evading}
    & 15.2 & 15.7 & 10.2 \\
    TI-MI-FGSM \cite{dong2019evading}
    & 22.8 & 24.6 & 14.8 \\
    TI-DIM \cite{dong2019evading}
    & 29.6 & 31.9 & 22.0 \\
    AdvRain-TI-MI-FGSM
    & \textbf{69.8} & \textbf{69.1} & \textbf{63.3} \\
    \bottomrule
    \end{tabular}
    }
  \end{minipage}
\end{table*}

\begin{table*}[t]
  \centering
  \begin{minipage}{\columnwidth}
    \newcommand{\tabincell}[2]{\begin{tabular}{@{}#1@{}}#2\end{tabular}}
    \caption{The mAP (in \%) of various adversarial attack methods against object detection on COCO.}
    \label{tab:Adv_OD_COCO_50}
    \centering
    \resizebox{0.99\linewidth}{!}{
    \begin{tabular}{lccccc}
    \toprule
    \multirow{2}{*}[-2pt]{\tabincell{c}{Attack\\Methods}} & \multicolumn{5}{c}{Threat Models (FR-v16)} \\
    \cmidrule(l){2-6}
    & FR-v16 & FR-mn & FR-rn50 & FR-rn101 & FR-rn152 \\
    \midrule
    Clean
    & 43.8 & 41.1 & 52.4 & 55.0 & 56.8 \\
    RAP \cite{li2018robust}
    &  3.3 &  9.8 & 24.6 & 29.8 & 22.3 \\
    UEA \cite{wei2019transferable}
    &  5.7 & 11.5 & 25.8 & 31.3 & 24.1 \\
    AdvRain
    & \textbf{2.2} & \textbf{8.3} & \textbf{21.5} & \textbf{27.5} & \textbf{19.7} \\
    \bottomrule
    \end{tabular}
    }
  \end{minipage}
  \hfill
  \begin{minipage}{\columnwidth}
    \newcommand{\tabincell}[2]{\begin{tabular}{@{}#1@{}}#2\end{tabular}}
    \caption{The mAP (in \%) of various adversarial attack methods against object detection on KITTI.}
    \label{tab:Adv_OD_KITTI_50}  
    \centering
    \resizebox{0.99\linewidth}{!}{
    \begin{tabular}{lccccc}
    \toprule
    \multirow{2}{*}[-2pt]{\tabincell{c}{Attack\\Methods}} & \multicolumn{5}{c}{Threat Models (FR-v16)} \\
    \cmidrule(l){2-6}
    & FR-v16 & FR-mn & FR-rn50 & FR-rn101 & FR-rn152 \\
    \midrule
    Clean
    & 74.6 & 72.3 & 75.8 & 78.1 & 78.4 \\
    RAP \cite{li2018robust}
    &  8.7 & 17.5 & 20.4 & 23.1 & 23.9 \\
    UEA \cite{wei2019transferable}
    & 10.3 & 21.6 & 23.2 & 25.0 & 26.3 \\
    AdvRain
    & \textbf{5.6} & \textbf{11.8} & \textbf{15.4} & \textbf{17.8} & \textbf{18.2} \\
    \bottomrule
    \end{tabular}
    }
  \end{minipage}
\end{table*}

\subsection{Evaluation of Rain Appearance}
\subsubsection{Visual Effect}
We compare the visual effect of our adversarial rainy images with that of other synthesized rainy images in Fig. \ref{fig:Eva_figs}. We observe that our adversarial rainy images show more realistic visual effects than the synthesized rainy images in databases Rain100H \cite{yang2017deep}, Rain800 \cite{zhang2019image}, Rain1200 \cite{zhang2018density} and Rain1400 \cite{fu2017removing}, which exhibit irregular rain directions or snow-like rain streaks.
In comparison with Fig. \ref{fig:Eva_figs} (c) and Fig. \ref{fig:Eva_figs} (h), the physics-based rain \cite{halder2019physics} produces relatively more realistic appearances than ours.
This is because the physics-based rainy images are synthesized by warping the rain streaks in a ready-made rain database \cite{garg2006photorealistic} with a physical particle system \cite{de2012fast}, and also use environmental lighting for rain rendering.
While our rainy images are generated directly from random noises, with which the adversarial ability should also be taken into consideration. We believe the visual effect of our adversarial rainy images can be further improved by employing lighting information, which we leave as future work.

\subsubsection{User study}
We also conduct a user study to further evaluate the rain appearance following \cite{halder2019physics}. Specifically, we randomly select 10 images from each rain database (\ie, our adversarial rainy images and other five synthesized rain databases), in collecting a total of 60 rainy images. Then, we invite 30 participants (12 undergraduate students, 18 graduate students; 13 female, 17 male; aged 20 to 34) to rate the visual effect of rainy images using a 5-points Likert scale. The results are summarized in Fig. \ref{fig:user_study}.
We can see that our adversarial rainy images and physics-based rainy images obtain similar high voting scores compared with other types of rainy images, confirming the visual realism of our rain generation method.

\subsection{Attacking Image Classification}
We report the comparison results of different adversarial attacks for image classification in Table \ref{tab:Adv_IC_DEV}. All adversarial attack methods are targeted at Inc-v3 (white-box attack), and also transferred to Inc-v4, IncRes-v2 and Xception (black-box attack).
For the white-box attack, the AdvRain achieves the highest success rate than all comparative methods. For the black-box attack, our AdvRain also has better transferability than conventional adversarial noise attacks.

We also compare our AdvRain with baselines on three defense models (Inc-v3$\rm{_{ens3}}$, Inc-v3$\rm{_{ens4}}$ and IncRes-v2$\rm{_{ens}}$) based on adversarial training from \cite{tramer2018ensemble}, and the results are listed in Table \ref{tab:Adv_defense}. We can see that our AdvRain still clearly outperforms its competitors in attacking adversarially trained models, which further demonstrates the advantages of our method.

\subsection{Attacking Object Detection}

Table \ref{tab:Adv_OD_COCO_50} and Table \ref{tab:Adv_OD_KITTI_50} summarize the performance of various adversarial attack methods against object detection on COCO and KITTI, respectively. For attack methods, we use FR-v16 as a targeted model, and then perform the transfer attack to FR-mn, FR-rn50, FR-rn101, and FR-rn152. For COCO and KITTI datasets, the performance is measured by mAP with IoU 0.7 and with the PASCAL criteria.
As we can see in the tables, the AdvRain consistently outperforms the RAP and UEA across all cases. Specifically, when attacking the white-box model FR-v16, the mAP score of FR-v16 drops significantly from 74.6\% to 5.6\%.

\begin{figure}[t]
  \centering
  \subfigure[]{\includegraphics[width=0.326\columnwidth]{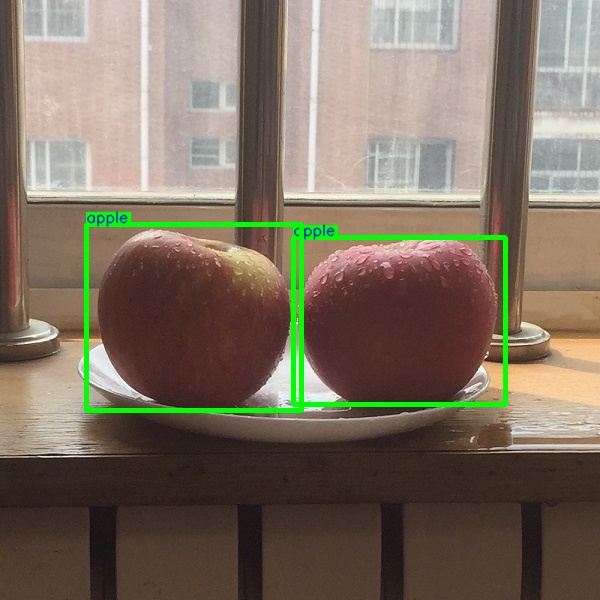}\label{fig:Real_attack_1}}
  \hfil
  \subfigure[]{\includegraphics[width=0.326\columnwidth]{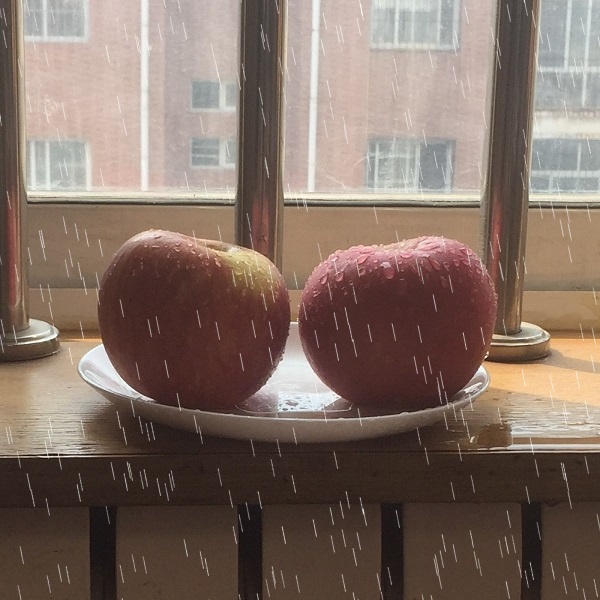}\label{fig:Real_attack_2}}
  \hfil
  \subfigure[]{\includegraphics[width=0.326\columnwidth]{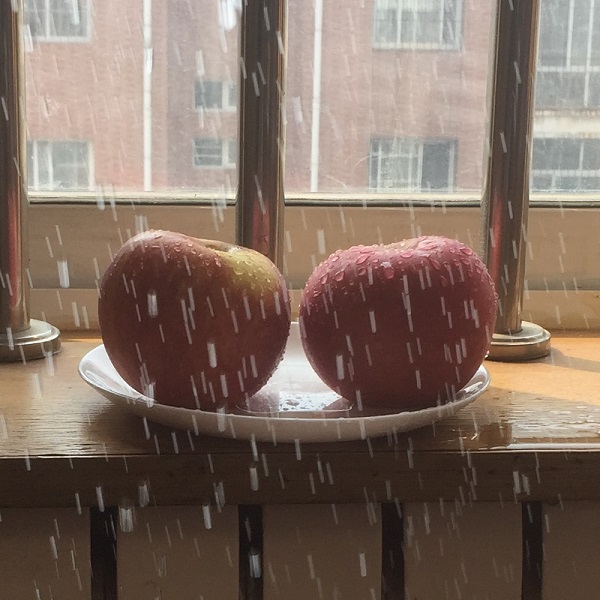}\label{fig:Real_attack_3}}
  \caption{Attack results in the real world. (a) Clean image. (b) AdvRain. (c) Real AdvRain.}
  \label{fig:Real_attack}
\end{figure}

\begin{figure*}[t]
	\centering
  	\includegraphics[width=1.99\columnwidth]{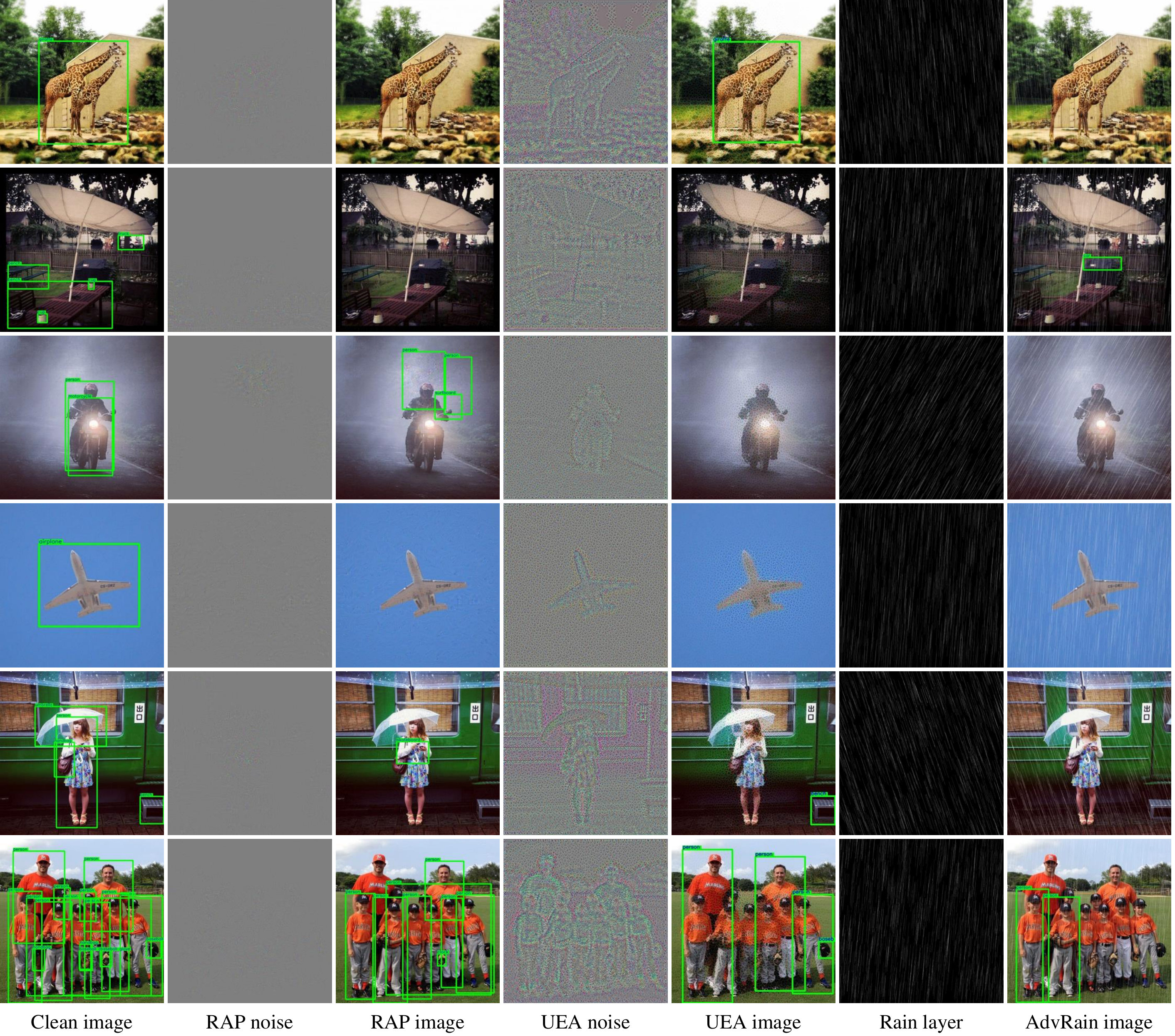}
	\caption{Qualitative comparisons between RAP \protect\cite{li2018robust}, UEA \protect\cite{wei2019transferable} and our AdvRain. Please zoom in to see the details.}
	\label{fig:visual_detect}
\end{figure*}

\subsection{Real-world Adversarial Rain Attack}
The adversarial rain attack produces adversarial examples by using the rain factors guided by DNNs. The rain factors, including rain noises $\mathbf N$, translation parameters $\theta$, and kernels ${\cal K}$, determine the adversarial capability of a specific rainy image. Therefore, a natural question arises, \ie, whether it is possible to use the obtained rain factors to adjust the appearance and dynamics of real rain so that to enable the rain to fool the DNNs. The resulting rain can be regarded as a type of real-world adversarial rain attack, which is performed in the following steps.

\begin{enumerate}
\item Fix the objects to be classified or detected by DNNs, and capture an image with a camera. 
\item Generate the corresponding rainy image with our adversarial rain attack, and also obtain the rain factor values.
\item Leverage the rain factors to adjust the simulated rain to achieve adversarial capability. \label{step3}
\end{enumerate}

For this experiment, we perform a physical behavior study in the real-world environment by our authors, using water-drops from cans to emulate the real rain effects.
In particular, we spray the water by a watering can so that the water drops are freely falling from a higher place. The desired rain intensity ($\mathbf N$) can be achieved by adjusting the amount of water drops, and the rain directions ($\theta$) are controlled by the wind speed and wind direction of an electric fan. The kernels ${\cal K}$ cannot be obtained in the real world, so we do not consider adjusting the rain textures.

We use the Faster-RCNN to detect the clean image, adversarial rainy image and real-world adversarial rainy image, the results of which are illustrated in Fig. \ref{fig:Real_attack}. 
We observe that both two types of adversarial rainy images can fool the object detector, demonstrating that the rain factors can be used as guidance for launching real-world adversarial attacks. The appearances of two adversarial rainy images are not very similar, but this problem can be adjusted by using more sophisticated rain simulation equipment.

Yet, one might argue that the real natural rain cannot be controlled with specific rain factors. Since the patterns and dynamics of natural rain are varied and diverse, the rain effect on DNNs is unpredictable. Our experiment is not to control the movement of natural rain, but yet to reveal the fact that the natural rain which thwarts the DNNs can be simulated by using the rain factors guided by the DNNs.

\begin{table}[t]
  \newcommand{\tabincell}[2]{\begin{tabular}{@{}#1@{}}#2\end{tabular}}
  \caption{Deraining performance in terms of PSNR and SSIM on Rain100H and Rain1400.}
  \label{tab:derainig}  
  \centering
  \resizebox{0.97\linewidth}{!}{
  \begin{tabular}{lcccc}
  \toprule
  \multirow{2}{*}[-2pt]{\tabincell{c}{Deraining Models}} & \multicolumn{2}{c}{Rain100H} &\multicolumn{2}{c}{Rain1400} \\
  \cmidrule(r){2-3} \cmidrule(l){4-5}
  & PSNR & SSIM & PSNR & SSIM \\
  \midrule
  PreNet \cite{ren2019progressive}
  & 29.44 & 0.8930 & 32.56 & 0.9452 \\
  PreNet-Flip-Rotation
  & 29.45 & 0.8931 & 32.58 & 0.9454 \\
  PreNet-AugMix
  & 30.10 & 0.8976 & 32.29 & 0.9468 \\
  PreNet-AdvRain
  & \textbf{30.65} & \textbf{0.9012} & \textbf{32.95} & \textbf{0.9480} \\
  \midrule
  EfDeRain \cite{guo2020efficientderain}
  & 30.35 & 0.8971 & 32.06 & 0.9248 \\
  EfDeRain-Flip-Rotation
  & 30.37 & 0.8976 & 32.10 & 0.9255 \\
  EfDeRain-AugMix
  & 30.62 & 0.9032 & 32.49 & 0.9278 \\
  EfDeRain-AdvRain
  & \textbf{31.04} & \textbf{0.9085} & \textbf{32.72} & \textbf{0.9303} \\
  \bottomrule
  \end{tabular}
  }
\end{table}

\begin{table}[t]
  \caption{The top 1 accuracy rates (in \%) of image classifiers on various types of images (clean images, rainy images and derained images).}
  \label{tab:defense}
  \centering
  \resizebox{0.99\linewidth}{!}{
  \begin{tabular}{lcccc}
  \toprule
  Deraining Models & Inc-v3 & Inc-v4 & IncRes-v2 & Xception \\
  \midrule
  Clean images     & 95.9 & 97.1 & 99.8 & 96.0 \\
  Rainy images     & 54.5 & 56.8 & 58.4 & 56.0 \\
  \midrule
  PreNet \cite{ren2019progressive}
  & 82.4 & 84.3 & 85.6 & 84.7 \\
  PreNet-AdvRain
  & \textbf{85.6} & \textbf{87.4} & \textbf{88.5} & \textbf{87.2} \\
  \midrule
  EfDeRain \cite{guo2020efficientderain}
  & 82.9 & 84.5 & 85.8 & 85.2 \\
  EfDeRain-AdvRain
  & \textbf{85.8} & \textbf{88.0} & \textbf{88.7} & \textbf{87.6} \\
  \bottomrule
  \end{tabular}
  }
\end{table}

\begin{figure*}[t]
  \centering
  \includegraphics[width=0.98\linewidth]{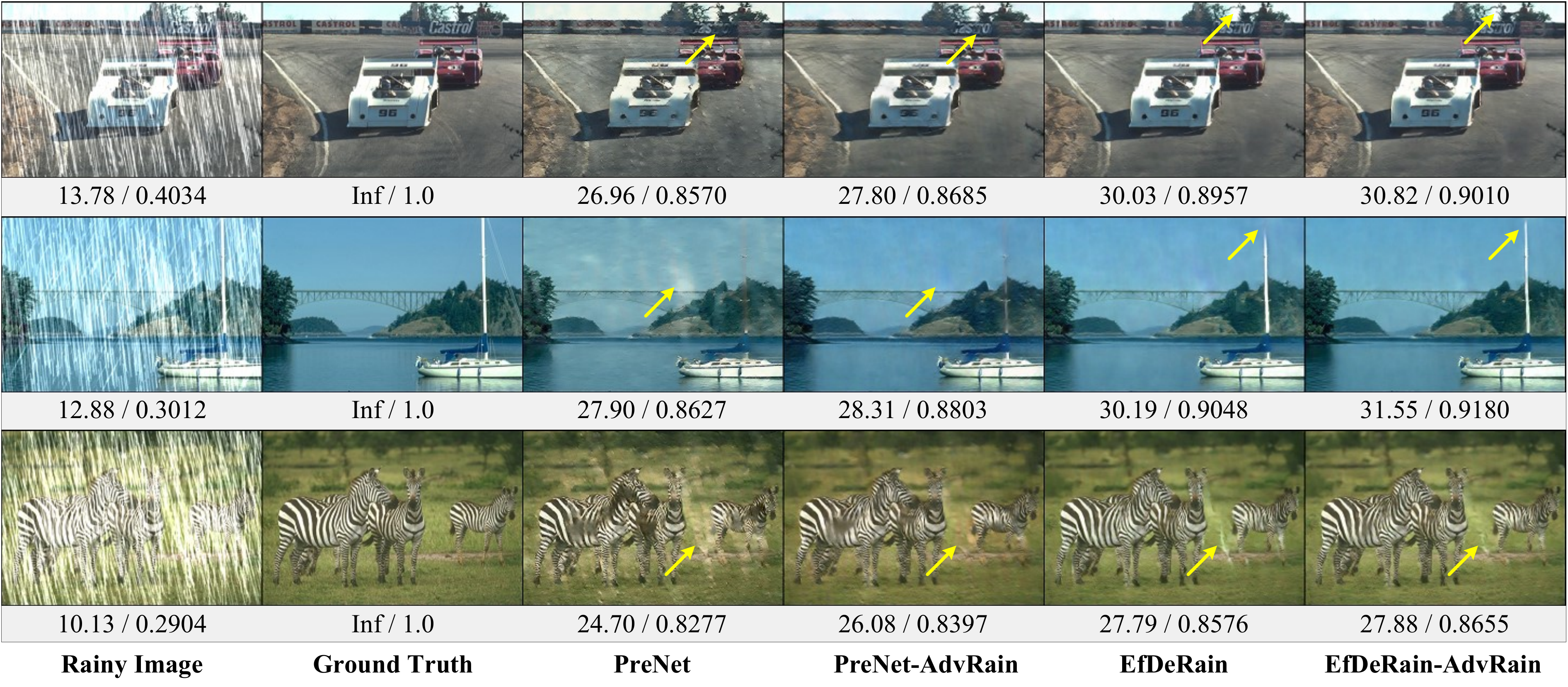}
  \caption{Visual comparison of deraining models with and without adversarial rain augmentation. The number below the figure indicates PSNR / SSIM values, and the yellow arrow in the figure shows the main difference between de-rained images.}
  \label{fig:derainig}
\end{figure*}

\subsection{Qualitative Comparisons of Adversarial Examples}\label{sec:qualitative}

We illustrate the qualitative comparisons between the adversarial examples of RAP \cite{li2018robust}, UEA \cite{wei2019transferable} and our AdvRain in Fig. \ref{fig:visual_detect}. For simplicity, we only compare the adversarial examples using FR-v16 (Faster RCNN with VGG16) \cite{ren2015faster} on COCO dataset \cite{lin2014microsoft} in a white-box attack scenario.

For the RAP, the adversarial noises are minimal perturbations which are mostly concentrated in the object areas to be detected. This type of small adversarial noises is determined by the perception constraints in terms of PSNR. However, the RAP still leaves perceptible artifacts appeared in the smooth texture images (see the third and the fourth rows in Fig. \ref{fig:visual_detect} for RAP), implying that the conventional adversarial noises are not applicable for all kinds of images. For the UEA, the adversarial noises are irregular latticed patterns, which are mainly located in object areas and also scattered throughout the images. The noise pattern in UEA is easily noticeable, and this is due to the generative mechanism of UEA which uses a pix2pix GAN \cite{isola2017image} to generate adversarial examples. Our adversarial rainy images are synthesized by compositing learned rain layers and images instead of directly adding noises to images, so they have high visual realism in spite of the visibility of rain.



\subsection{Evaluation of Defensive Deraining} \label{exp_deraining}
In this subsection, we first evaluate the effectiveness of adversarial rain augmentation on training deraining models. Besides, to explore how the deraining models trained with our adversarial rain augmentation could improve the performance of DNN perception, we evaluate the classification accuracy on rainy images with and without deraining.

For the deraining task, we use PreNet \cite{ren2019progressive} and EfDeRain \cite{guo2020efficientderain} as baselines, and report the deraining results on dataset Rain100H \protect\cite{yang2017deep} and Rain1400 \protect\cite{fu2017removing} in Table \ref{tab:derainig}. To verify the superiority of our adversarial  rain  augmentation , we also use two conventional data augmentation methods for comparison, one is AugMix \cite{hendrycks2019augmix} using multiple normal rainy layers, and the other one is flip and rotation using a single normal rainy layer.

We can see from Table \ref{tab:derainig} that for Flip\&Rotation and AugMix, the improvement of deraining performance is quite marginal by just adding more synthesized rainy images into training. This is because the training data not only need rainy images with more diverse rain situations, but also need to cover the tough and formidable rainy cases which can more easily fool the DNNs. The latter requirement can only be satisfied by our AdvRain. In contrast, the deraining models trained with our adversarial rain (indicated by suffix AdvRain) consistently achieves substantial PSNR and SSIM gains over the corresponding baseline models, confirming that the adversarial rain does simulate more rainy patterns and provide more hard rain examples, thus improving the deraining performance. Furthermore, we also provide some visual examples from dataset Rain100H \protect\cite{yang2017deep} in Fig. \ref{fig:derainig}, from which we can see that the deraining models with adversarial rain augmentation produce more clear images with more detail patterns.

Having demonstrated the advantage of our adversarial rain augmentation on deraining, we further use the trained deraining models to improve the performance of image classification by removing the rain streaks in images before they are fed to classifiers. We use the images in DEV dataset as clean images, and use the NormalRain2 in Section \ref{advrain_normalrain} as rainy images. We utilize four image classifiers (Inc-v3, Inc-v4, IncRes-v2 and Xception) to evaluate the effectiveness of deraining, and report the top 1 accuracy rates in Table \ref{tab:defense}, in which the results in the last four rows are obtained from the derained images with different deraining models.

We observe from Table \ref{tab:defense} that the image classifiers all achieve high accuracy rates on clean images, but have low accuracy rates on rainy images. After removing the rain streaks in the rainy images, the image classifiers significantly increase the accuracy rates. Most notably, the derained images using our augmented deraining model are more easily to be identified by image classifiers, indicating that our defensive deraining with  adversarial rain augmentation is indeed beneficial to the DNN perception systems.

\section{Conclusion}\label{sec:concl}

In this paper, we comprehensively study the potential risks of rain effect on DNN perception. We propose a new type of adversarial rain attack, which simulates natural rain situations with the guidance of DNNs and thus synthesizes visually realistic rainy images to mislead image classification and object detection. The adversarial rain attack reveals the essential and inevitable threat factors of rain, which commonly exist in the real world.
We also study the elimination of of rain effect on DNNs, and propose a defensive deraining strategy that leverages adversarial rain augmentation to enhance the deraining models for downstream perception tasks.
We conducted extensive experiments to validate the effectiveness of our approaches on different datasets and tasks.
Our results show that the current state-of-the-art DNNs can be vulnerable to the inevitable rain effects in the real-world, and the proposed adversarial rain can also be used to augment the rainy image databases for deraining tasks, all of which call the attention to take rain effects into consideration for more advanced design of real-world DNN-based perception systems.

In future work, we would like to further optimize our adversarial rain by considering lighting conditions and fog-like rains, and also extend the adversarial rain attack to a more broad range of applications, such as semantic segmentation \cite{lateef2019survey} and visual object tracking \cite{fiaz2019handcrafted}.
\ifCLASSOPTIONcaptionsoff
  \newpage
\fi



%



\bibliographystyle{IEEEtran}
\bibliography{ref}

%








\end{document}